\documentclass{mic2011}

\usepackage[ruled,vlined,commentsnumbered]{algorithm2e}
\usepackage{pstricks}
\usepackage{pst-node}

\begin{document}

\title{Neigborhood Selection in Variable Neighborhood Search}
\author{M. J. Geiger\inst{1} \and M. Sevaux\inst{1,2} \and Stefan Vo\ss{}\inst{3}}
\institute{
  Helmut Schmidt University,
    Logistics Management Department\\
    Holstenhofweg 85, 22043 Hamburg, Germany\\
  \email{m.j.geiger@hsu-hh.de}
  \and
  Universit\'e de Bretagne-Sud,
  Lab-STICC -- Centre de Recherche\\
  2 rue de St Maud\'e, F-56321 Lorient, France\\
  \email{marc.sevaux@univ-ubs.fr}
  \and
  University of Hamburg,
  Institute of Information Systems (Wirtschaftsinformatik),\\
  Von-Melle-Park 5, 20146 Hamburg, Germany\\
  \email{stefan.voss@uni-hamburg.de}
}
\id{S2-15}
\maketitle

\section{Introduction}

Variable neighborhood search (VNS)
\cite{hansen.mladenovic:99,hansen.mladenovic:03}
is a metaheuristic for solving optimization problems based on a
simple principle: systematic changes of neighborhoods within the
search, both in the descent to local minima and in the escape from
the valleys which contain them.
VNS applications have been numerous and successful. Many
extensions have been made, mainly to be able to solve large
problem instances. Nevertheless, a main driver behind VNS is to
keep the simplicity of the basic scheme. Variable neighborhood
descend (VND) belongs to the family of methods within VNS. The
idea behind VND is to systematically switch between different
neighborhoods where a local search is comprehensively performed
within one neighborhood until no further improvements are
possible. Given the found local optimum, VND continues with a
local search within the next neighborhood. If an improved solution
is found one may resort to the first neighborhood; otherwise the
search continues with the next neighborhood, etc.

Designing these neighborhoods and applying them in a meaningful
fashion is not an easy task. Moreover, incorporating these
neighborhoods, e.g., into VNS may still need a considerable degree
of ingenuity and may be confronted with the following research
questions (among others):

\vspace{-2ex}
\begin{itemize}\addtolength{\itemsep}{-0.6\baselineskip}
  \item Which neighborhoods can be designed?
  \item How should they be explored and how efficient are they?
  \item How structurally different are they?
  \item In which order should they be applied?
\end{itemize}

In this paper we attempt to investigate especially the latter
concern. Assume that we are given an optimization problem that is
intended to be solved by applying the VNS scheme, how many and
which types of neighborhoods should be investigated and what could
be appropriate selection criteria to apply these neighborhoods.
More specifically, does it pay to ``look ahead'' (see, e.g.,
\cite{HoMeVo2008} in the context of VNS and GRASP) when attempting
to switch from one neighborhood to another? Is it reasonable to
apply ``nested neighborhood structures'' (like first 2-optimal
exchanges, then 3-optimal exchanges, then 4-optimal etc.) or
should neighborhood structures be considerably different from each
other? Often different neighborhoods are proposed allowing for
ideas regarding sequential and nested changes.

We cannot provide a comprehensive survey on the VNS in this
extended abstract. We emphasize just very few earlier works on VNS
investigating some of the questions raised.

For instance, in \cite{Burke201046} VNS is hybridized with a
genetic algorithm (GA). Chromosomes in the GA represent the
possible choices of neighborhoods to be applied. However, as the
authors state, ``the ordering of neighborhoods is unimportant
since VNS cycles through all neighborhoods.'' On the other hand,
when hybridizing the VNS with the pilot method as it is performed
in \cite{HoMeVo2008} it actually seems favorable to put some
effort in an appropriate choice of the next neighborhood. Problems
considered are exam timetabling in the first case and
telecommunications network design in the latter. These results are
in line with the observations in \cite{HuRaidl06} for applying VND
on a different network design problem.

To summarize, different sources come to different conclusions
making it even more important to investigate these topics further.
While our research questions are of generic nature we attempt to
investigate them and provide answers with respect to the
\textit{single machine total weighted tardiness problem} (SMTWTP)
as a specific combinatorial optimization problem. Though this
setting might not necessarily lead to generalizable results (as is
the case with the references cited above) they seem to allow
additional insights into the VNS method in itself.

\section{VNS for the Single Machine Total Weighted Tardiness Problem}

The SMTWTP or $1|\ |\sum w_jT_j$ is a well-known machine
scheduling problem. In the SMTWTP, a set of $n$ jobs with weights
indicating their relative importance needs to be processed on a
single machine while minimizing the total weighted tardiness. For
a recent paper on the SMTWTP plus some related references see,
e.g., \cite{geiger:10}.

With respect to our above mentioned research questions, the SMTWTP
may be seen as a representative problem with the following
characteristics: easy to represent by some straightforward
encoding, hard to solve, an existing testbed of problem instances
with known optimal solutions is readily available for detailed
analysis.

\paragraph{Neighborhoods for the SMTWTP.}

The set of neighborhoods to be explored should be important enough
to get some insights regarding our purpose. We have decided to
build a set of neighborhoods composed by the classical ones and by
some structurally different neighborhoods.

 The investigated neighborhoods include the following:

\vspace{-2ex}
\begin{description}\addtolength{\itemsep}{-0.6\baselineskip}
  \item[APEX] it is the most classical neighborhood for permutation encodings. Two adjacent jobs are exchanged in the sequence. For a solution, the size of this neighborhood is $n-1$.
  \item[BR4] consists in taking a block of four consecutive jobs and inverse its internal orientation. The size of this neighborhood is $n-3$.
  \item[BR5] is identical to BR4 but with a block of five jobs. The size of this neighborhood is $n-4$.
  \item[BR6] is identical to BR4 but with a block of six jobs. The size of this neighborhood is $n-5$.
  \item[EX$\setminus$APEX] is the general pairwise interchange where the APEX is excluded. Two non-adjacent jobs are exchanged. The size of this neighborhood is $n(n-3)+2$.
  \item[FSH$\setminus$APEX] takes a job and moves it to a position further in the sequence, resulting in a forward shift for the jobs in between these two positions.
        APEX is excluded from this move. The size of this neighborhood is $\sum_{i=1}^{n-2}i=(n-2)(n-1)/2$.
  \item[BSH$\setminus$APEX] works as FSH$\setminus$APEX but the jobs are shifted backward. The size is identical to FSH$\setminus$APEX.
\end{description}


\paragraph{Experimental Design.}

Since we want to observe the behavior of the search when selecting
the different neighborhoods, we propose three different
approaches. To simplify the experiments and not be too dependent
of randomness, instead of using a VNS, we first use a VND scheme.

In the experiments, we will compare the three following algorithms:

\vspace{-2ex}
\begin{description}\addtolength{\itemsep}{-0.6\baselineskip}
  \item[VND-R] When a local optimum is reached, the next neighborhood is selected randomly among possible alternatives.
        The search terminates when all the neighborhoods cannot produce a better solution than the incumbent.
  \item[VND-F] This is the classical version of VND. In our experiments,
        we choose the order in which neighborhoods are introduced above.
  \item[VND-A] This is an adaptive version of VND. Every time a new neighborhood has to be selected,
        an investigation of all neighborhoods is run for a short number of iterations (say, 100) and the neighborhood producing
        the best solution at this time is selected for the descent method. This neighborhood is used until no more improvements are obtained.
        The investigation phase is run again, etc. The stopping condition is the same as before.
\end{description}

To make a fair comparison, for all approaches we count the number
of evaluations made during the searches and we shall use graphs
indicating the quality of the best solution \emph{vs.} the number
of evaluations. To consider nested neighborhoods, the above
exclusion may be abolished, too.

\section{Preliminary results}

\begin{figure}[htbp]
  \centering
  \scalebox{.55}{ 
  \psset{unit=0.9cm}
  \begin{pspicture*}(-0.5,0.2)(10.5,8)
    \rput(5,4){\includegraphics[angle=-90,width=10cm]{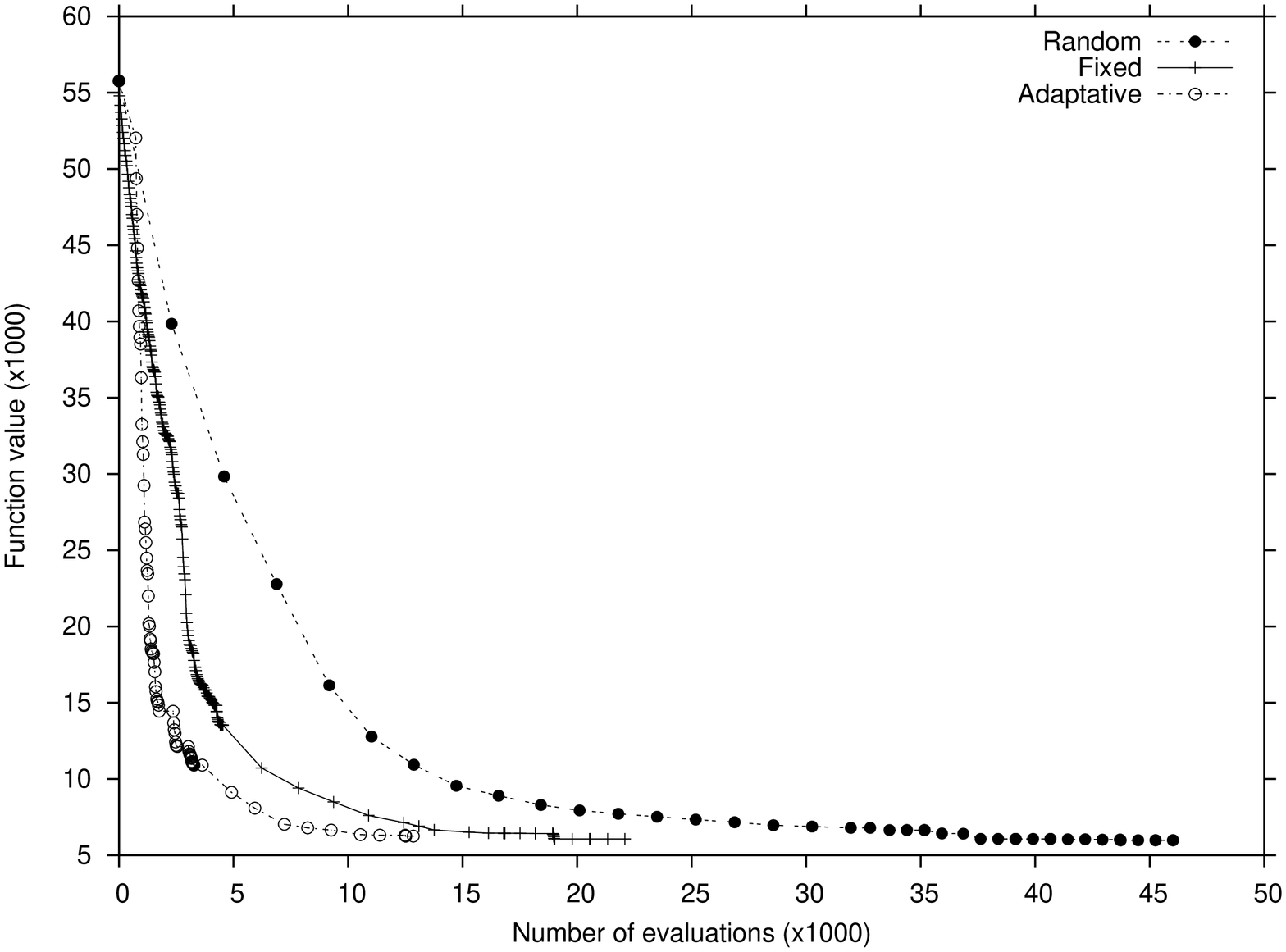}}
    \Cnode(3.15,1.05){ea} \Cnode(4.9,1.05){ef} \Cnode(9.39,1.05){er}
    \rput[B](7,3){\rnode{ea2}{\sffamily\scriptsize End of VND-A}}
    \rput[B](7,2.5){\rnode{ef2}{\sffamily\scriptsize End of VND-F}}
    \rput[B](7,2){\rnode{er2}{\sffamily\scriptsize End of VND-R}}
    \ncdiagg[angleA=180,arm=.5,nodesepA=3pt]{->}{ea2}{ea}
    \ncdiagg[angleA=180,arm=.5,nodesepA=3pt]{->}{ef2}{ef}
    \ncdiagg[angleA=0,arm=.5,nodesepA=3pt]{->}{er2}{er}
    \rput[B](5,5.5){\rnode{x1}{\sffamily\scriptsize VND-A improves VND-R}}
    \rput[B](5,5.0){\rnode{x2}{\sffamily\scriptsize VND-A improves VND-F}}
    \pnode(.9,6.62){c1} \pnode(.9,5.6){c2}
    \ncdiagg[angleA=180,arm=.5,nodesepA=3pt]{->}{x1}{c1}
    \ncdiagg[angleA=180,arm=.5,nodesepA=3pt]{->}{x2}{c2}
  \end{pspicture*}
  }
  \caption{Example run for the first instance of \cite{crauwels.potts.ea:98}}
  \label{fig:first}
\end{figure}

Experiments are conducted on the classical 100 job instances from
\cite{crauwels.potts.ea:98}. These instances have some
advantages: best known solutions are stable (not improved for a
long time), the number of instances is not too large (125) and
these instances still seem difficult to solve (despite their
relatively small size with 100 jobs).
 In Figure \ref{fig:first} we provide as an example the results of
the first run on the first of the investigated set of problem
instances. In this case we find that VND-R takes longer to stop
than VND-F, and that takes more time than VND-A. That is, we may
deduce a positive aspect of the adaptive version of VND. We can
also see that if we run VND for less than 750 evaluations, VND-R
gives best results, and less than 834 evaluations give priority to
VND-F. Of course, 834 evaluations are not enough to reach good
quality and VND-A shows its strength here.

This ideal situation was unfortunately not observed as a general
behavior and we report considerably more details in the final
version of the paper together with related conclusions and
proposals for related adaptivity in VNS and VND with the final
conclusion that adaptivity actually pays.

\bibliography{mic-geiger-etal}
\bibliographystyle{plain}

\end{document}